\documentclass[letterpaper]{article} % DO NOT CHANGE THIS
\usepackage{aaai2026}  % DO NOT CHANGE THIS
\usepackage{times}  % DO NOT CHANGE THIS
\usepackage{helvet}  % DO NOT CHANGE THIS
\usepackage{courier}  % DO NOT CHANGE THIS
\usepackage[hyphens]{url}  % DO NOT CHANGE THIS
\usepackage{graphicx} % DO NOT CHANGE THIS
\usepackage{natbib}  % DO NOT CHANGE THIS AND DO NOT ADD ANY OPTIONS TO IT
\usepackage{caption} % DO NOT CHANGE THIS AND DO NOT ADD ANY OPTIONS TO IT
\usepackage{algorithm}
\usepackage{algorithmic}

\usepackage{newfloat}
\usepackage{listings}
\DeclareCaptionStyle{ruled}{labelfont=normalfont,labelsep=colon,strut=off} % DO NOT CHANGE THIS
\floatstyle{ruled}
\newfloat{listing}{tb}{lst}{}
\floatname{listing}{Listing}
\usepackage{array}
\usepackage{tabularx}
\usepackage{booktabs}
\usepackage{amsmath}
\usepackage{amsfonts}
\usepackage{pdfpages}

\title{Geometry Meets Light: Leveraging Geometric Priors for Universal Photometric Stereo under Limited Multi-Illumination Cues}
\author {
    King-Man Tam\textsuperscript{\rm 1},
    Satoshi Ikehata\textsuperscript{\rm 2,3},
    Yuta Asano\textsuperscript{\rm 2},
    Zhaoyi An\textsuperscript{\rm 1},
    Rei Kawakami\textsuperscript{\rm 1}
}
\affiliations {
    \textsuperscript{\rm 1}Institute of Science Tokyo\\
    \textsuperscript{\rm 2}National Institute of Informatics\\
    \textsuperscript{\rm 3}Denso IT Laboratory\\
    tam.k.4f46@m.isct.ac.jp,
    sikehata@nii.ac.jp
}

\usepackage{bibentry}
\begin{document}

\maketitle

\begin{abstract}
Universal Photometric Stereo is a promising approach for recovering surface normals without strict lighting assumptions. However, it struggles when multi-illumination cues are unreliable, such as under biased lighting or in shadows or self-occluded regions of complex in-the-wild scenes. We propose GeoUniPS, a universal photometric stereo network that integrates synthetic supervision with high-level geometric priors from large-scale 3D reconstruction models pretrained on massive in-the-wild data. Our key insight is that these 3D reconstruction models serve as visual-geometry foundation models, inherently encoding rich geometric knowledge of real scenes. To leverage this, we design a Light-Geometry Dual-Branch Encoder that extracts both multi-illumination cues and geometric priors from the frozen 3D reconstruction model. We also address the limitations of the conventional orthographic projection assumption by introducing the PS-Perp dataset with realistic perspective projection to enable learning of spatially varying view directions. Extensive experiments demonstrate that GeoUniPS delivers state-of-the-arts performance across multiple datasets, both quantitatively and qualitatively, especially in the complex in-the-wild scenes.
\end{abstract}

\begin{links}
    \link{Code}{https://github.com/marcotam2002/geounips}
\end{links}

\section{Introduction}

Photometric Stereo (PS)~\cite{woodham1980photometric} is a method for recovering high-fidelity surface normals from multiple images captured under varying illumination with a fixed camera. Historically, the development of PS can be seen as a gradual relaxation of assumptions about the lighting conditions. Traditional PS methods relied on physically based inverse rendering with calibrated directional lighting and specific BRDFs (e.g., Lambertian)~\cite{Ikehata2012,Shi2014}. Early learning‑based methods showed that normals of non‑Lambertian, non‑convex surfaces could be directly regressed from images in an uncalibrated setup, but they still assumed directional lighting models~\cite{Chen2020,Sarno2022}.
More recently, physics‑free universal PS methods~\cite{Ikehata2022,Ikehata2023,Ikehata2024} have removed the need for explicit lighting models, enabling use with arbitrary uncalibrated light sources.

\begin{figure}
    \centering
    \includegraphics[width=\linewidth]{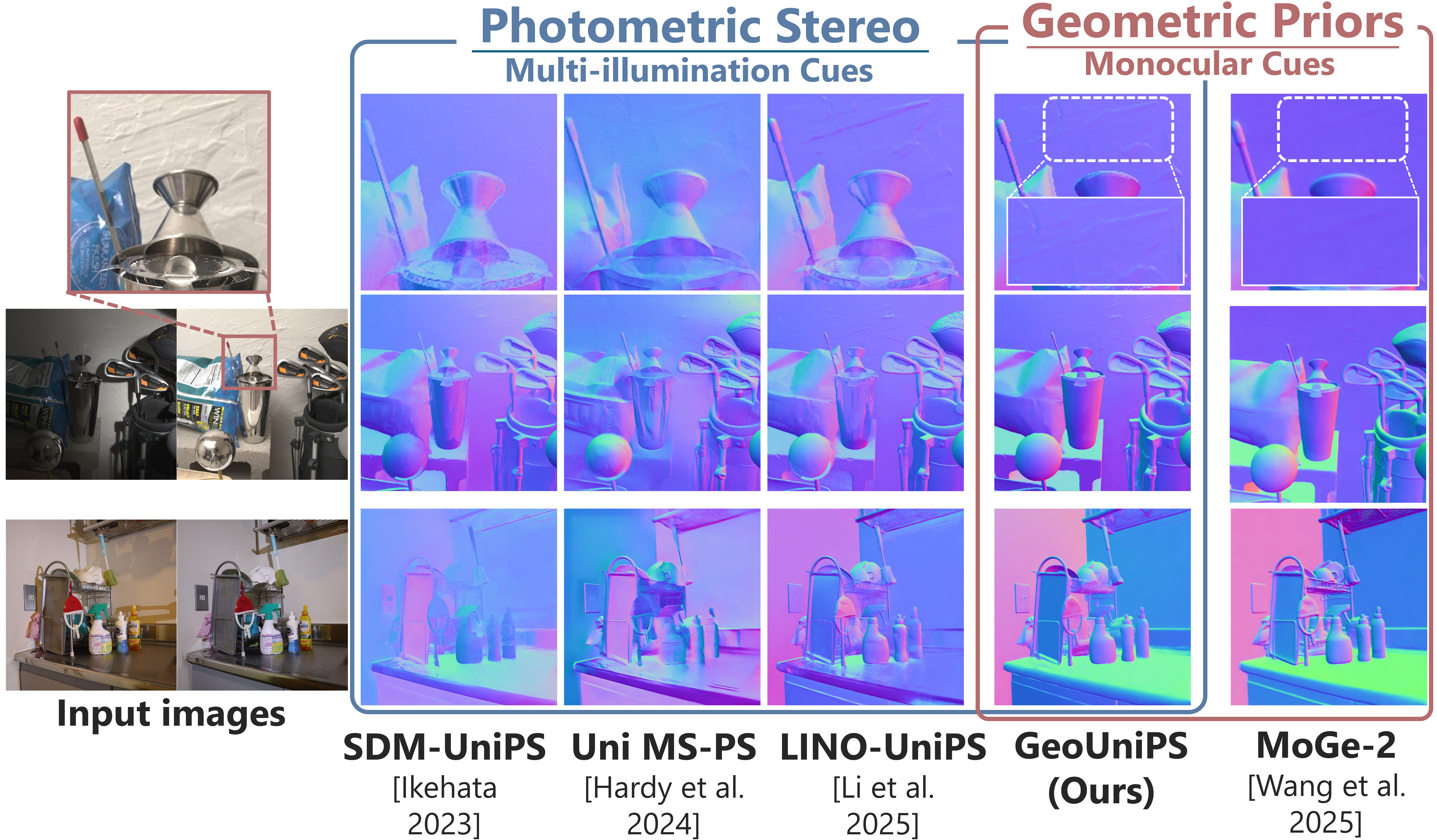}
    \caption{Our method effectively leverages geometric priors from pretrained 3D reconstruction model, achieving more plausible normal map recovery in challenging scenes with complex backgrounds and limited lighting variation. Compared to SoTA monocular normal prediction models (e.g., MoGe-2~\cite{wang2025moge2}), our approach captures finer surface details by incorporating multi-illumination cues.}
    \label{fig:teaser}
\end{figure}

While recent advances have eliminated many assumptions about lighting, one critical premise remains: {\it each surface point is assumed to be observed under sufficiently diverse, well-distributed lighting}. However, this condition often fails in real-world settings. Due to the practical difficulty of controlling light sources, some regions may receive rich and varied illumination, while others receive far less. As illustrated in Fig.~\ref{fig:teaser} (bottom), the scene was illuminated using a moving handheld flashlight from the front. Performance in poorly illuminated areas degrades significantly, as also discussed in~\cite{Ikehata2023}; the degradation is even more pronounced in regions with complex geometry (e.g., containers), material (e.g., mirror surface) and textures (e.g., bottles), where limited lighting cues make normal estimation more difficult. This limitation fundamentally arises because photometric stereo relies on multi-illumination cues induced by changes in illumination as its primary cue; when these variations are unreliable, the method lacks a mechanism to compensate.

A straightforward way to handle this can be to train models on large‑scale in‑the‑wild multi-illumination datasets to learn monocular geometric priors, following recent 3D reconstruction studies. For example, recent feedforward multiview 3D reconstruction models trained on millions of in‑the‑wild images, e.g., ~\cite{Wang2024dust3r, wang2025vggt}, have shown that even with a single input image, a network can recover a plausible 3D shape, despite being trained with multiple images, indicating that these models have learned high‑level monocular priors beyond low‑level multiview photometric constraints.

However, applying this strategy to photometric stereo is extremely challenging. Acquiring ground‑truth normal maps for real scenes with the resolution and fidelity required by PS is prohibitively expensive, and it is even more difficult to cover the vast combinations of lighting conditions, surface properties and camera settings. Consequently, existing PS models are  exclusively trained on clean synthetic datasets, whose statistics differ markedly from those of real scenes which naturally encode rich geometric and contextual priors (e.g., building façades tend to be piecewise planar). Without exposure to such real‑world regularities, PS models could largely rely on shading variations, leaving poorly illuminated regions without meaningful guidance. This raises a crucial question: {\it How can we acquire high‑level geometric priors in a synthetic photometric stereo training pipeline?}

In this paper, we present {\it GeoUniPS}, a geometry‑aware universal photometric stereo network that combines synthetic supervision with high‑level priors from large‑scale 3D reconstruction models (e.g., VGGT~\cite{wang2025vggt}). Our key insight is that 3D reconstruction models, pretrained on massive in‑the‑wild datasets, act as visual‑geometry foundation models and inherently encode rich geometric knowledge of real scenes. We find that by injecting their features into the photometric stereo pipeline, models can leverage geometric priors unattainable through purely synthetic multi‑illumination training. Even with limited illumination variation, these priors regularize surface normal estimation, yielding more reliable and globally consistent results as illustrated in Fig.~\ref{fig:teaser} (top).

Technically, we propose a novel dual-branch encoder: one branch extracts lighting-aware features through synthetic supervision, while the other captures lighting-invariant, high-level geometric features from a frozen 3D reconstruction model. These complementary cues are fused into a unified representation, which the decoder leverages to produce context-aware, geometry-faithful normals. Unlike prior methods that rely solely on multi-illumination cues, GeoUniPS benefits from the pretrained model’s embedded geometric priors, enabling contextually valid estimations even when multi-illumination cues are limited.

We further address a gap in existing training datasets. Previous photometric stereo datasets typically assume orthographic projection, whereas real world setups operate under perspective projection. To bridge this gap, we construct PS-Perp, the first synthetic training dataset with realistic perspective projection, enabling the network to learn spatially varying view directions.

In our evaluation, we demonstrate that GeoUniPS enriched with priors from 3D reconstruction models, achieves more plausible results across various datasets. These include standard photometric stereo benchmarks under single directional lighting~\cite{boxin2016diligent,luces2021}, as well as real world multi-illumination datasets~\cite{multiill19}.

\section{Related Work}
\subsubsection{Photometric Stereo:}
Photometric stereo has a long history in computer vision, estimating surface normals of static scenes from images under varying illumination. The classic method~\cite{woodham1980photometric} assumes Lambertian, convex surfaces, and known directional lighting in a darkroom. Later works extended it to handle non-Lambertian scenes, using robust techniques that treat non-Lambertian effects as outliers~\cite{Wu2010,Ikehata2012} or explicitly use non-Lambertian BRDF~\cite{Alldrin2007b,Goldman2010}.

With the progress of deep learning, data-driven photometric stereo emerged under calibrated lighting, targeting non-Lambertian, non-convex surfaces via observation maps~\cite{Ikehata2018}, set-pooling~\cite{Chen2018}, graph neural networks~\cite{Yao2020}, and Transformers~\cite{Ikehata2021}. Meanwhile, neural inverse rendering methods~\cite{taniai2018neural,Li2022a} adopted physics-guided, unsupervised learning to estimate normals without supervision. Uncalibrated settings were also addressed by self-calibrating networks predicting light and normals sequentially~\cite{chen2019self,Chen2020,Kaya2021}.

However, these methods assume simplified lighting models, limiting their applicability under complex real-world lighting conditions. To address this, universal photometric stereo~\cite{Ikehata2022} has been proposed, aiming to learn lighting representations directly from images without restrictive lighting assumptions. SDM-UniPS~\cite{Ikehata2023} eliminated the need for masks and enabled high-resolution normal recovery using a pixel-sampling Transformer. Uni MS-PS~\cite{hardy2024uni} extended this approach to a multi-scale architecture, while SpectraM-PS~\cite{Ikehata2024} introduced the first universal multispectral photometric stereo networks for dynamic surfaces. Most recently, LINO-UniPS~\cite{li2025light} decoupled lighting from geometry, enhancing detail preservation through wavelet-based processing and a gradient-aware loss.

Despite recent progress, photometric stereo remains challenged by limited or biased lighting, where multi-illumination cues are not reliable. Its reliance on synthetic training data, which lacks the geometric context of real-world scenes, further limits generalization. These issues highlight the need for techniques to induce stronger, high-level geometric priors.

\begin{figure*}[t]
    \centering
    \includegraphics[width=\linewidth]{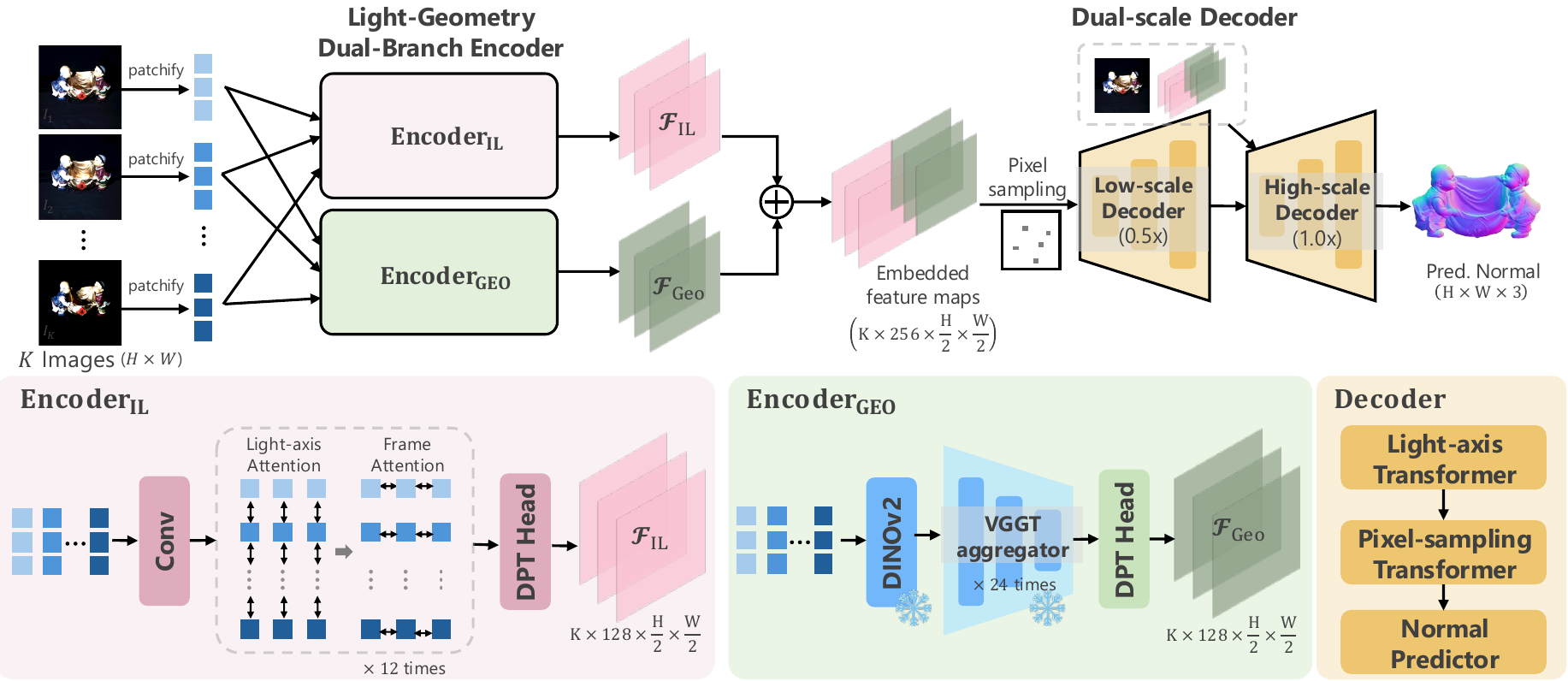}
    \caption{Overview of our GeoUniPS architecture. Given multiple input images captured under different lighting conditions, the Light-Geometry Dual-Branch Encoder extracts both light-variant features from multi-illumination cues ($\text{Encoder}_\text{IL}$) and geometric features from the pretrained VGGT aggregator ($\text{Encoder}_\text{Geo}$). These features are concatenated with the input images using an MLP-based embedding, after which the Dual-Scale Normal Decoder performs pixel-wise normal regression at sampled locations.}
    \label{fig:enter-label}
\end{figure*}

\subsubsection{Feedfoward 3D Reconstruciton Models:}
Recent feed-forward 3D reconstruction models have shifted multi-view reconstruction from traditional SfM + MVS pipelines to end-to-end neural inference. DUSt3R~\cite{Wang2024dust3r} demonstrates that a Transformer-based model can reconstruct dense point maps from unposed image pairs in a single pass. Building on this, MASt3R~\cite{Leroy2024mast3r} improves dense correspondence quality in wide-baseline scenarios. Multi-view extensions such as VGGT~\cite{wang2025vggt} and MV-DUSt3R+~\cite{Tang2025mvdust3r} further scale this paradigm. For instance, VGGT can process anywhere from a single image to hundreds of views in under a second, predicting camera intrinsics/extrinsics, depth maps, point maps, and even point tracks in a feed-forward manner. Crucially, VGGT shows that even a single image can suffice to reconstruct dense geometry, confirming that these networks encode high-level geometric knowledge beyond simple matching cues. Trained on large-scale in-the-wild datasets such as MegaDepth, CO3D-v2, ScanNet, and DL3DV, these models generalize robustly and demonstrate zero-shot performance in novel and challenging scenarios. This provides strong evidence that feed-forward 3D reconstruction models have learned rich priors about scene geometry directly from large-scale data.

\section{Method}
We propose \textbf{GeoUniPS}, the first universal photometric stereo network that integrates geometric priors from pretrained 3D reconstruction models with multi-illumination cues. We begin by formulating the universal photometric stereo problem and algorithm, then describe our architecture and training dataset design. We address two main challenges in real-world scenario: (1) limited illumination cues, such as weak illumination variations, shadows, or self-occlusion, which degrade surface normal estimation; and (2) the orthographic projection assumption, which fails to adapt to perspective scenes in real-world settings. To overcome them, we propose both architectural and data-driven solutions.

On the architectural side, we introduce a \textit{Light-Geometry Dual-Branch Encoder} that extracts both high-level geometric priors and multi-illumination cues. Unlike previous methods that rely solely on multi-illumination cues, we incorporate geometric guidance from a pretrained foundation model (e.g., VGGT~\cite{wang2025vggt}) to compensate for limited multi-illumination cues.

To adapt our model to general perspective scenes, we first demonstrate the universal photometric stereo networks trained on a large scale synthetic dataset rendered under \textit{perspective projection}, comprising over 60{,}000 scenes with diverse shapes, materials, and focal lengths. This dataset exposes the model to realistic perspective distortions.
\subsection{Problem Statement and Algorithm Formulation}

Universal photometric stereo~\cite{Ikehata2022} is a variant of the photometric stereo task where no information about the lighting conditions is available (i.e., no calibration nor prior lighting models), unlike conventional calibrated~\cite{woodham1980photometric} or uncalibrated~\cite{Chen2019} PS tasks. Specifically, given a set of $K$ images $\mathcal{I}=\{I_k\}_{k=1}^K$, \[I_k \in \mathbb{R}^{H \times W \times 3}\], captured by a fixed camera under arbitrary lighting conditions (commonly referred to as multi-illumination images), the goal of universal photometric stereo is to recover a pixel-accurate surface normal map $N \in \mathbb{R}^{H \times W \times 3}$. Importantly, the number of images $K$ is arbitrary at test time. An optional object mask $M \in \mathbb{R}^{H \times W}$ is often provided, but our method does not assume the availability of this mask at inference (yet used during training), as it introduces additional cost.

Most existing universal photometric stereo networks~\cite{Ikehata2022,Ikehata2023,Ikehata2024,li2025light} are characterized by a two-stage design: an encoder that extracts image-level feature maps from individual images with intra- and inter- frame interactions, and a decoder that recovers the surface normal at randomly sampled pixel locations using the extracted features and the original image information. Following them, GeoUniPS also has the same two-stage design.

The role of the encoder is to extract $K$ feature maps from $\mathcal{I}$. Formally, \begin{equation}
\mathcal{F}=\{F_k\}_{k=1}^K = \text{Encoder}(\mathcal{I}), \quad F_k \in \mathbb{R}^{H' \times W' \times C},
\end{equation}
where $C$ is the feature dimension, and $H' \times W'$ is the feature map size, typically smaller than $H \times W$.

Given $P$ spatial coordinates $\mathcal{X} = \{\boldsymbol{x}_p\}_{p=1}^P$ where $\boldsymbol{x}_p \in [1, H] \times [1, W]$, the decoder predicts the surface normal $\boldsymbol{n}_p \in \mathbb{R}^3$ at each position $\boldsymbol{x}_p$ from feature maps and images. Formally, 
\begin{equation}
\{\boldsymbol{n}_p\}_{p=1}^P = \text{Decoder}(\mathcal{F}(\mathcal{X}\mathord{\downarrow}), \mathcal{I}(\mathcal{X})),
\end{equation}
where $\mathcal{X}\mathord{\downarrow}$ represents the downsampled coordinates corresponding to the resolution change from $H \times W$ to $H' \times W'$. The encoder and decoder are trained solely with normal vector supervision on a large-scale synthetic dataset.

This two-stage design is inspired by traditional photometric stereo, where each image is paired with lighting information (e.g., light direction vectors), and surface normals are estimated by aggregating these pairs per pixel. In universal photometric stereo networks, this role is replaced by features extracted from images. The encoder’s feature maps thus serve as substitutes for lighting; SDM-UniPS~\cite{Ikehata2023} refers to them as the Global Lighting Context (GLC).

Predicting normals at sparse pixels reduces memory and aligns with the core idea of estimating per-pixel normals from illumination variation without spatial context. Some variants incorporate spatial cues, such as patch-level features~\cite{Ikehata2024} or coarse-to-fine strategies~\cite{hardy2024uni}.

To isolate the contribution of geometric priors from pretrained models, {\it we follow this simple and established framework, minimizing confounding factors and ensuring that observed improvements are attributable to the geometric priors themselves}. The following sections detail the encoder and decoder design in GeoUniPS.
\subsection{Light-Geometry Dual-Branch Encoder}
We design our encoder to capture both illumination-aware and illumination-invariant priors. Specifically, the encoder consists of two branches:

\begin{itemize}
\item \textbf{$\text{Encoder}_{\text{Geo}}$}: This branch leverages high-level geometric priors that are invariant to illumination. We use the encoder from a pretrained 3D reconstruction model trained on large-scale in-the-wild data, followed by a learnable projector that adapts its output to our task. The pretrained encoder is frozen during training to preserve its geometric knowledge, which is expected to complement the illumination priors.
\item \textbf{$\text{Encoder}_{\text{IL}}$}: This branch captures a multi-illumination prior by embedding shading variations across images taken under different lighting conditions from a fixed viewpoint. It follows the motivation of conventional universal photometric stereo networks. This information is essential for the decoder to recover fine-grained surface normals from multi-illumination images.
\end{itemize}

The combined feature representation is obtained by simply concatenating the outputs of the two branches:
\begin{equation}
\mathcal{F} = \text{Concat}\left(\text{Encoder}_{\text{Geo}}(\mathcal{I}), \text{Encoder}_{\text{IL}}(\mathcal{I})
\right),
\end{equation}
where $\text{Concat}(\cdot)$ denotes channel-wise concatenation.

\textbf{$\text{Encoder}_{\text{Geo}}$} consists of a frozen encoder of a pretrained 3D reconstruction model, followed by a learnable projector. We adopt VGGT~\cite{wang2025vggt} for its strong zero-shot geometry reconstruction capabilities. The projector is implemented as a DPT-head~\cite{Ranftl2021dpt}. Although VGGT was originally designed for multi-view reconstruction, it has been shown to perform well even with a single image by leveraging strong monocular geometric priors. 

GeoUniPS uses VGGT's aggregator without its decoders. As shown in Fig.~2, each input image is resized or cropped to a resolution divisible by 14, in accordance with VGGT’s architectural constraints. After preprocessing (i.e., DINOv2 normalization) and tokenization via DINOv2, the image tokens are passed to VGGT's aggregator, which comprises 24 layers of alternating frame and global attention. We extract tokens from layers [4, 11, 17, 23] and fuse them using a learnable DPT head, projecting the VGGT features into a 128-dim feature map at a 2× downsampled resolution, denoted as \( \mathcal{F}_{\text{Geo}} \in \mathbb{R}^{K\times \frac{H}{2} \times \frac{W}{2} \times 128} \).

\textbf{$\text{Encoder}_{\text{IL}}$} adopts a Transformer-based architecture similar to VGGT’s aggregator with a DPT head, but replaces the DINOv2 tokenizer with a lightweight convolution layer to better capture fine-grained local patterns. The number of layers is reduced, as this encoder learns priors limited to smaller synthetic datasets. All parameters are trained from scratch.

Inter-image interaction is key to extracting illumination-aware features in this branch. Following prior universal photometric stereo methods, we replace VGGT’s full attention with light-axis attention. Note that, removing inter-image attention from $\text{Encoder}_{\text{Geo}}$ showed negligible effect, likely because VGGT’s aggregator extracts strong geometric priors even from multi-illumination images as the view is fixed.

As in $\text{Encoder}_{\text{Geo}}$, each input is resized or cropped. After max-val normalization (per SDM-UniPS~\cite{Ikehata2023}), the convolutional tokenizer processes images into tokens. These are passed to a modified VGGT's aggregator with $12$ layers, alternating between frame attention (intra-image) and light-axis attention (inter-image at the same spatial position). This design enables joint reasoning over spatial context and illumination differences. We extract tokens from layers [2, 5, 8, 11] and fuse them via a DPT head into a 128-dim feature map at $2\times$ downsampled resolution, denoted as \( \mathcal{F}_{\text{IL}} \in \mathbb{R}^{K\times \frac{H}{2} \times \frac{W}{2} \times 128} \).

Finally, the embedded feature maps are given as
\begin{equation}
\mathcal{F} = \text{Concat}\left(\mathcal{F}_{\text{Geo}}, \mathcal{F}_{\text{IL}}\right).
\end{equation}

\subsection{Pixel-Sampling-Based Normal Decoder}

While our primary focus is on the encoder, which leverages pretrained 3D reconstruction models, an effective normal decoder is also essential for recovering detailed surface normals. We explore several decoder variants, including different embeddings of $\mathcal{I}$ (i.e., pixel~\cite{Ikehata2023}, patch~\cite{Ikehata2024}, and MLP-based~\cite{li2025light}) and architectural designs (single-scale~\cite{Ikehata2023,li2025light} vs.\ dual-scale~\cite{Ikehata2024}). Although these variants achieve comparable benchmark scores, we find that the dual-scale decoder with MLP-based embedding provides the best trade-off between geometric detail and efficiency. Therefore, we adopt this design for our decoder.

$\text{Decoder}_{\text{GeoUniPS}}(\mathcal{F}(\mathcal{X}{\downarrow}), \mathcal{I}(\mathcal{X}))$ predicts surface normals at randomly sampled pixel locations $\mathcal{X}$ from extracted features $\mathcal{F}$ and input images $\mathcal{I}$. Henceforth, we treat the features at $\mathcal{X}$ as tokens and apply Transformer layers to them. As shown in SDM-UniPS, the number of sampled pixels directly influences normal estimation accuracy, with larger sample sizes generally yielding better results. To balance performance with training efficiency, we sample 2,048 pixels during training and increase this to 10,000 pixels during inference. This sampling size can be further raised when memory permits, potentially leading to improved accuracy.

First, low-scale normals are estimated from $\mathcal{F}(\mathcal{X}{\downarrow})$ using five 256-dimensional light-axis Transformers, followed by a 384-dimensional light-axis Transformer with Pooling-by-Multihead-Attention (PMA)~\cite{Lee2019} to aggregate features along the light-axis. Two 384-dimensional pixel-sampling Transformers~\cite{Ikehata2023} then apply self-attention across spatial locations to enhance spatial coherence. Finally, a two-layer MLP (384$\rightarrow$192$\rightarrow$3) predicts the low-frequency normals, which are normalized to unit length.

For high-scale refinement, $\mathcal{I}(\mathcal{X})$ is passed through a 256-dimensional MLP with two LayerNorms (3$\rightarrow$256), embedding RGB values into a high-dimensional space. The embedded features are concatenated with $\mathcal{F}(\mathcal{X}{\downarrow})$, processed by five 256-dimensional light-axis Transformers (512$\rightarrow$256), and aggregated along the light-axis via a 384-dimensional PMA (256$\rightarrow$384). These features are fused with the low-scale normals into 387-dimensional representations. Spatial coherence is further enhanced by two 384-dimensional pixel-sampling Transformers, followed by a final MLP (387$\rightarrow$384$\rightarrow$192$\rightarrow$3) that predicts the normals, which are normalized to unit length. The final normal map is reconstructed by aggregating predictions over all locations.
\\
\noindent \textbf{Training Loss:}
The training loss was computed using the Mean Squared Error (MSE) loss function to measure the $\ell_2$ error between the predicted and ground truth surface normal vectors. This loss was calculated at both scales and then summed.
\subsection{PS-Perp: Perspective Synthetic Training Dataset}
To improve generalization to real-world scenes captured by perspective cameras, we introduce PS-Perp, the first large-scale synthetic dataset for Universal Photometric Stereo constructed using a perspective projection model. While existing datasets such as PS-Wild~\cite{Ikehata2022}, PS-Mix~\cite{Ikehata2023}, LINO-UNIPS~\cite{li2025light}, and others~\cite{hardy2024uni, Yamaguchi_2025_WACV} have contributed variations in scene and material complexity, they are all rendered under the orthographic camera assumption. This limits their ability to train models that generalize to scenes exhibiting strong perspective distortion.

Unlike prior work, PS-Perp is rendered with a perspective camera using Blender’s Cycles renderer. Focal lengths are sampled from a broad range (20–1000mm) to simulate varying levels of distortion: shorter focal lengths (e.g., 20–70mm) induce strong perspective effects, while longer ones (e.g., 70-1000mm) approximate weak perspectives. In total, the dataset consists of 60,297 multi-object scenes, with 44,220 scenes rendered using focal lengths below 70mm to emphasize perspective diversity, and the remaining using longer focal lengths to include weakly distorted cases. Each scene is rendered into 10 16-bit images at 512×512 resolution under randomized combinations of directional, point, and environment lighting, following the PS-Mix~\cite{Ikehata2023} pipeline. By sharing the same asset library, scene composition strategy, and lighting setup as PS-Mix, our dataset ensures compatibility while significantly expanding the coverage of camera models. PS-Perp enables training on a continuum from strongly to weakly distorted perspective images, bridging the gap between synthetic training data and realistic test conditions.

Since perspective cameras with long focal lengths cannot fully replicate orthographic views, we train our model using a combination of PS-Perp and PS-Mix~\cite{Ikehata2023} to cover both projection types. Some samples of training data are provided in the appendix.

\section{Results}\label{sec:ablation}

\subsection{Implementation Details and Computational Time}
As described, our network was trained from scratch on a combination of PS-Perp and PS-Mix~\cite{Ikehata2023}. The model is implemented in PyTorch and trained on 4 NVIDIA H100 GPUs over 6 days using the AdamW~\cite{Adam} optimizer, with an initial learning rate of 1e-4, a weight decay of 0.05, and step decay (multiplied by 0.8 every 10 epochs). A linear warm-up is applied during the first epoch. For numerical stability, we use full-precision (FP32) training. Each batch contains 2 scenes, with a randomly sampled number of input images ranging from 3 to 6 to improve robustness under varying lighting conditions. We evaluate accuracy using the Mean Angular Error (MAE) between the predicted and ground-truth normals. On a single H100 GPU, inference takes approximately 13 seconds for 16 images at a resolution of 512×512, excluding I/O.

\subsection{Evaluation Dataset}
We mainly used three public datasets for evaluation. For quantitative analysis, the following two object-centric datasets were used: \textbf{DiLiGenT}~\cite{boxin2016diligent}, which contains 10 objects captured under 96 directional lights with orthographic projection and provides 16-bit images, lighting information, masks, and ground truth normals; and \textbf{LUCES}~\cite{luces2021}, which includes 14 objects imaged under 52 near-field lights using a calibrated perspective camera (12-bit RAW), with full lighting parameters and ground truth normals and depths from 3D scans. Note that we don't use lighting information. 

For the qualitative analysis with more challenging scenes, we use \textbf{Multi-illumination dataset}~\cite{multiill19}. This dataset comprises 1,016 HDR indoor scenes captured under 25 directional lighting conditions using bounced flash illumination. Unlike object-centric datasets with uniform lighting, this scene-level dataset features spatially varying light distributions, where the geometry and materials of each room cause illumination to differ significantly across the scene. Direct lighting is limited to only 7 out of 25 directions; the rest produce purely indirect light via ceiling and wall bounces, yielding realistic, diverse lighting effects across entire environments. No normal map is provided.

We compared our method with SDM-UniPS~\cite{Ikehata2023}, Uni-MSPS~\cite{hardy2024uni}, and LINO-UniPS~\cite{li2025light}, which were introduced in the related work section.

\begin{figure*}[h]
    \centering
    \includegraphics[width=150mm]{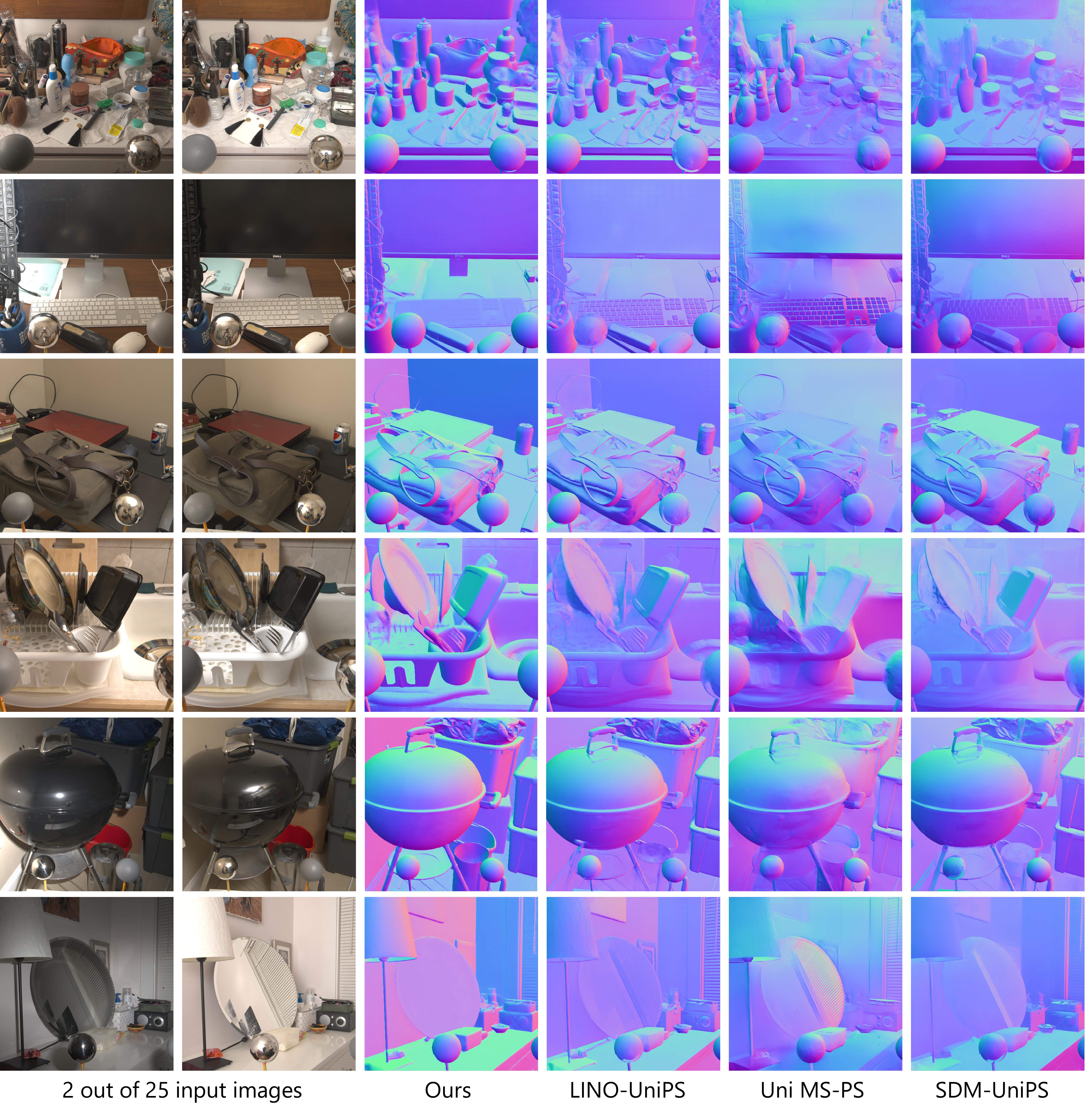}
    \caption{Qualitative comparison on Multi-illumination Dataset~\cite{multiill19}.}
    \label{fig:multi_illumination}
\end{figure*}

\subsection{Evaluation on DiLiGenT and LUCES}
We evaluate our method on the DiLiGenT~\cite{boxin2016diligent} (orthographic) and LUCES~\cite{luces2021} (perspecive) benchmarks under directional lighting. The purpose of this experiment is to validate that our method performs well on standard universal photometric stereo tasks, and that its effectiveness is not compromised by the introduction of geometric priors. Note that the primary goal is to demonstrate the availability of geometric priors from 3D reconstruction models when multi-illumination cues are unreliable, rather than to achieve the best performance on well-illuminated datasets such as DiLiGenT and LUCES.

Nevertheless, as shown in Table~\ref{table:diligent_eval} and~\ref{table:luces_eval}, our method achieves state-of-the-art performance on both datasets. Due to space constraints, all recovered normal maps are provided in the appendix. Notably, incorporating geometric priors significantly improves performance when multi-illumination cues are limited ($K=1$), enabling robust predictions with fewer input images. However, the method still performs best when sufficient illumination cues are available. For LUCES, while LINO-UniPS achieves competitive accuracy through wavelet-based refinement, our model demonstrates the comparable performance without the wavelet-based refinement.

Interestingly, we observe that using fewer input images sometimes yields better results. We attribute this to training with 3–6 input views, where evaluating on larger sets may lead to performance drops due to distribution shift.
\subsection{Evaluation on Multi-illumination Dataset}
We conducted a qualitative comparison using the Multi-illumination Dataset~\cite{multiill19}. Seven scenes were randomly selected for the main evaluation, with additional results provided in the Appendix. As ground-truth surface normals are not available for this dataset, our analysis is limited to qualitative assessment; nonetheless, several noteworthy observations emerged.

As shown in Fig.~\ref{fig:multi_illumination}, this dataset poses significantly greater challenges than conventional object-centric datasets. Unlike previous datasets with limited lighting directions and simpler scenes, it features complex geometry, greater depth variation, and more realistic spatial layouts. A key finding is that our method clearly benefits from the geometric priors learned via the pretrained 3D reconstruction model (i.e., VGGT). In particular, it captures plausible normals on floors, walls, and regions with complex depth variation or mirror/transparent surfaces—areas where methods relying solely on multi-illumination cues tend to fail.

Importantly, our method does not depend exclusively on geometric priors. The high-frequency components of the predicted normal maps are comparable to those of other methods, indicating that our approach effectively captures fine surface details from multi-illumination cues. This is especially evident in comparisons with MoGe~\cite{wang2025moge}, a monocular depth estimation method, as shown in Fig.~\ref{fig:teaser}.
\subsection{Analytical Studies}

\setlength{\tabcolsep}{4pt} % 將欄間距從預設6pt減為4pt
\begin{table}[b]
\centering
\footnotesize
\caption{Evaluation of SDM-UniPS trained on different training datasets. MAE~↓ under varying focal lengths (mm).}
\begin{tabularx}{\linewidth}{@{}l*{5}{>{\centering\arraybackslash}X}@{}}
\toprule
\textbf{Training Dataset} & \textbf{15\,mm} & \textbf{35\,mm} & \textbf{70\,mm} & \textbf{200\,mm} & \textbf{Ortho} \\
\midrule
PS-Mix (Orthographic)      & 22.18 & 14.09 & 10.36 & 8.89  & 5.52 \\
PS-Perp (Perspective)      & 7.18  & 5.47  & 5.38  & 5.75  & 8.95 \\
PS-Perp + PS-Mix           & 6.98  & 5.53  & 5.53  & 5.90  & 5.62 \\
\bottomrule
\end{tabularx}
\label{tab:ablation_focal}
\end{table}

\begin{table*}[t]
\centering
\footnotesize
\caption[\hspace{1em}Evaluation on DiLiGenT (Mean Angular Errors in degrees). All 96 images were used except where K is shown.]{Evaluation on DiLiGenT (Mean Angular Errors in degrees). All 96 images were used except where K is shown.}

\begin{tabular}{cccccccccccc}
\toprule
\textbf{Method} & Ball & Bear & Buddha & Cat & Cow & Goblet & Harvest & Pot1 & Pot2 & Reading & Avg. \\
\midrule
SDM-UniPS \cite{Ikehata2023} & \textbf{1.45} & 3.50 & 7.54 & 5.19 & 4.48 & 7.69 & 10.76 & 4.59 & 4.41 & 8.43 & 5.80 \\
Uni MS-PS (Hardy et al.\ \citeyear{hardy2024uni}) & 1.93 & 3.56 & 6.53 & 4.13 & 4.12 & 7.35 & 9.88 & 4.31 & 4.49 & 7.45 & 5.38 \\
LINO-UniPS \cite{li2025light} & 1.77 & 2.62 & 6.22 & 3.38 & 4.38 & 5.14 & 8.60 & 4.07 & 4.54 & \textbf{6.75} & 4.75 \\
Ours & 2.63 & 2.46 & \textbf{5.95} & \textbf{3.27} & 3.93 & \textbf{5.00} & \textbf{8.54} & \textbf{3.81} & \textbf{4.00} & 6.88 & \textbf{4.65} \\
\midrule
Ours (K=16) & 2.75 & \textbf{2.45} & 6.04 & 3.44 & \textbf{3.84} & \textbf{5.00} & 8.88 & 3.96 & 4.24 & 7.12 & 4.77 \\
Ours (K=4)  & 3.49 & 2.73 & 7.66 & 5.58 & 4.09 & 5.63 & 11.15 & 4.62 & 5.43 & 8.76 & 5.91 \\
Ours (K=1) & 6.70 & 5.74 & 17.26 & 12.94 & 10.78 & 10.26 & 26.53 & 9.72 & 10.06 & 18.60 & 12.86 \\
\bottomrule
\end{tabular}
\label{table:diligent_eval}
\end{table*}

\begin{table*}[t]
\centering
\footnotesize
\caption[\hspace{1em}Evaluation on LUCES (Mean Angular Errors in degrees). All 52 images were used except where K is shown.]{Evaluation on LUCES (Mean Angular Errors in degrees). All 52 images were used except where K is shown.}
\setlength{\tabcolsep}{2pt} % 控制基本欄距
\begin{tabular*}{\textwidth}{@{\extracolsep{\fill}}lccccccccccccccc}
\toprule
\textbf{Method} & Ball & Bell & Bowl & Buddha & Bunny & Cup & Die & Hippo & House & Jar & Owl & Queen & Squirrel & Tool & Avg \\
\midrule
SDM-UniPS       & 11.77 & 12.92 &  8.66 & 18.16 &  8.83 & 11.36 &  7.22 &  8.95 & 25.91 &  8.84 & 12.82 & 15.30 & 15.92 & 12.58 & 12.80 \\

Uni MS-PS       & 11.62 & 11.66 &  7.96 & 13.38 & 10.02 & \textbf{7.92} &  6.50 &  8.80 & 25.62 &  6.35 & 12.07 & 12.77 & 12.18 & 11.24 & 11.29 \\

LINO-UniPS      &  9.65 & \textbf{8.97} &  8.26 & 13.30 &  5.67 &  8.30 &  6.25 &  5.82 & 22.69 & \textbf{6.13} & \textbf{9.29} & \textbf{9.98} & 10.56 & \textbf{7.55} & 9.46 \\

Ours            &  7.59 & 10.22 &  8.13 & 13.11 &  5.50 & 10.05 &  3.79 & \textbf{5.62} & \textbf{21.84} &  6.17 & 10.76 & 10.56 & \textbf{9.86} &  8.61 & \textbf{9.42} \\
\midrule
Ours (K=16)     & \textbf{7.42} & 10.71 &  7.83 & \textbf{13.10} & \textbf{5.35} &  9.80 & \textbf{3.71} &  5.63 & 22.00 &  6.43 & 10.56 & 10.79 & 11.17 &  8.65 &  9.62 \\
Ours (K=4)      &  7.75 & 12.22 & \textbf{7.46} & 14.25 &  5.39 &  8.57 &  4.06 &  6.15 & 24.14 &  6.85 & 12.29 & 12.85 & 13.27 &  8.52 & 10.27 \\
Ours (K=1)      &  9.08 & 10.12 & 11.78 & 15.71 & 11.14 & 15.05 &  7.31 & 12.31 & 32.71 &  8.80 & 16.11 & 20.67 & 21.60 & 10.78 & 14.33 \\
\bottomrule
\end{tabular*}
\label{table:luces_eval}
\end{table*}
\paragraph{Effect of Training Data.}
To assess the benefit of our training dataset design independently of our architecture, we train SDM-UniPS~\cite{Ikehata2023} from scratch using three different datasets: (1) PS-Mix~\cite{Ikehata2023}, rendered under orthographic projection; (2) PS-Perp, our proposed dataset with varying focal lengths under perspective projection; and (3) a hybrid combination of both. These three models are evaluated on 100 newly rendered samples, generated in the same manner as PS-Mix and PS-Perp but with a different focal length. All models were trained for three days using identical hyperparameters.

As shown in Table~\ref{tab:ablation_focal}, the model trained solely on PS-Mix performs poorly under strong perspective distortion, with an MAE of 22.18° at 15mm, although performance improves as the focal length increases. In contrast, the model trained on PS-Perp generalizes well to perspective images (e.g., 7.18° at 15mm) but degrades when applied to orthographic images. The hybrid training consistently performs well across all focal lengths, validating the complementary nature of the two datasets.

\paragraph{Effect of Encoder Design.}
We conduct an analytical study to evaluate the encoder design by comparing three configurations: (1) $\text{Encoder}_{\text{Geo}}$ only, (2) $\text{Encoder}_{\text{IL}}$ only, and (3) our full Light-Geometry Dual-Branch Encoder. To isolate encoder effects from decoder influence, we use the single-scale+pixel-embedding decoder from SDM-UniPS~\cite{Ikehata2023}. To test generality of our method beyond VGGT~\cite{wang2025vggt}, we also implement $\text{Encoder}_{\text{Geo}}$ with MoGe~\cite{wang2025moge}, a monocular depth estimator trained on large-scale in-the-wild datasets. All models are trained on the combined PS-Perp and PS-Mix datasets for about 400,000 iterations. Evaluation is conducted on DiLiGenT~\cite{boxin2016diligent} (orthographic) and LUCES~\cite{luces2021} (perspective) benchmarks using MAE over all objects.

As shown in Table~\ref{table:encoder}, $\text{Encoder}{\text{IL}}$ only degrades significantly when multi-illumination cues are absent ($K=1$), resulting in poor normal estimation. With larger $K$, these cues become more reliable, improving performance. In contrast, $\text{Encoder}{\text{Geo}}$ only does not scale well with more input images, despite using illumination cues in the decoder, highlighting the importance of extracting them in the encoder. Our Dual-Branch Encoder, while slightly worse than $\text{Encoder}_{\text{IL}}$ only on DiLiGenT with $K=4$ and $K=16$, performs robustly across both $K=1$ and $K=16$. It notably outperforms others on LUCES, likely because geometric priors are more effective under perspective projection. Comparing backbones, VGGT consistently outperforms MoGe, validating our choice. However, both show similar trends, confirming that geometric priors benefit PS tasks regardless of the pretrained model.

\begin{table}[t]

\centering
\footnotesize
\caption{Ablation study on encoder configurations. MAE~(↓) for different K on DiLiGenT ~\cite{boxin2016diligent} and LUCES ~\cite{luces2021}. All models use a single decoder.}
\label{tab:ablation_encoder}
\setlength{\tabcolsep}{3pt} % 壓縮欄距
\begin{tabular}{lcccccc}
\toprule
Encoder & \multicolumn{3}{c}{DiLiGenT} & \multicolumn{3}{c}{LUCES} \\
 & K=1 & K=4 & K=16 & K=1 & K=4 & K=16 \\
\midrule
$\text{Encoder}_{\text{Geo}}$ w/ MoGe      & 13.63 & 8.96 & 8.33 & 15.50 & 12.32 & 11.01 \\
$\text{Encoder}_{\text{Geo}}$ w/ VGGT      & 13.07 & 8.27 & 6.90 & 14.70 & 11.90 & 11.12 \\
% Light (Axis) Encoder & 20.84 & 9.60 & 7.25 & 21.05 & 13.91 & 12.36 \\
$\text{Encoder}_{\text{IL}}$  & 19.03 & 6.50 & 4.96 & 19.82 & 11.40 & 10.36 \\
Dual-Branch Encoder    & 12.84 & 6.81 & 5.19 & 14.82 & 10.60 & 9.82 \\
\bottomrule
\end{tabular}
\label{table:encoder}
\end{table}

\paragraph{Comparison with 3D Scan.}
While the Multi-illumination Dataset highlights the advantage of our method under limited lighting diversity, its lack of ground-truth normals and single-directional lighting may raise concerns. To address this, we conduct a quantitative evaluation using four images captured under weak, non-uniform lighting from a selfie ring light with ambient illumination. Lighting movement was minimized to suppress cues, and object masks were not provided. This setup is intentionally difficult and serves as an analytical stress test rather than a realistic scenario.

\begin{figure}
    \centering
    \includegraphics[width=\linewidth]{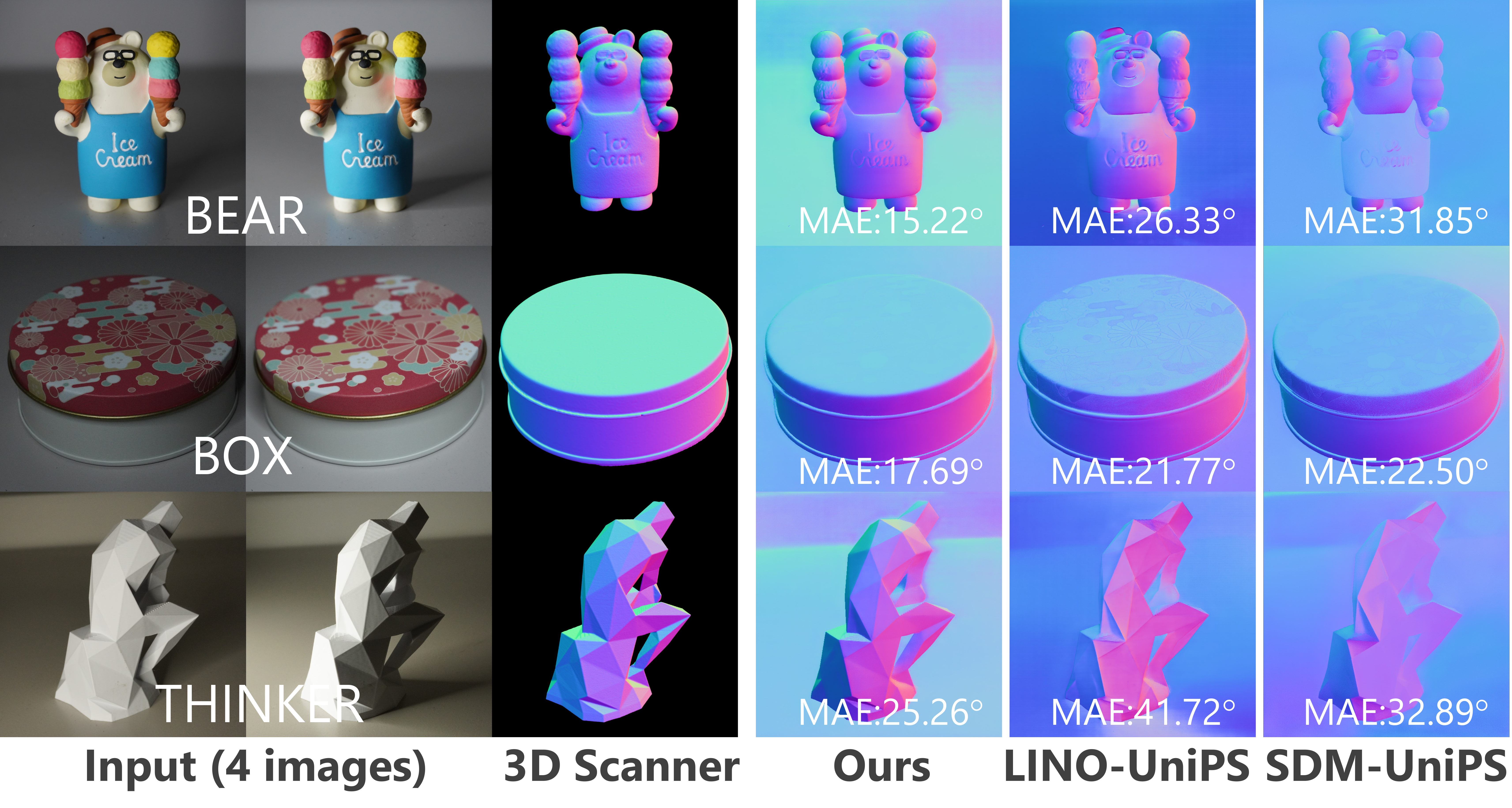}
    \caption{Results for scenes under limited lighting cues, excluding object masks.}
    \label{fig:marco}
\end{figure}

3D scans were obtained using an EinScan-SE scanner and aligned for evaluation. As shown in Fig.~\ref{fig:marco} the results, both SDM-UniPS and LINO-UniPS fail almost completely, while our method, though not perfect, performs significantly better. This suggests that leveraging pretrained geometric knowledge is highly effective, especially under conditions where traditional photometric stereo struggles.
\section{Conclusion}
We introduced GeoUniPS, a geometry-aware universal photometric stereo network designed to address the limitations of existing methods under insufficient or biased illumination. Motivated by the observation that real-world scenes often lack diverse lighting cues, we proposed leveraging high-level geometric priors from pretrained 3D reconstruction models. To this end, we designed a Light-Geometry Dual-Branch Encoder that jointly captures lighting-aware features and lighting-invariant geometric features. GeoUniPS achieves state-of-the-art performance on both orthographic and perspective benchmarks and demonstrates strong qualitative results on complex real-world scenes. 

While we leverage 3D reconstruction models as geometric priors, exploring alternative backbones such as diffusion-based geometry estimators or other large-scale pretrained models (e.g., segmentation models), remains future work. Additionally, the performance boost under sufficient lighting cues is limited, likely due to our simple feature concatenation. Designing better fusion strategies is an important direction ahead.
\section{Aknowledement}
This work was supported by JSPS KAKENHI Grant Number 24K02966 and DENSO IT LAB Recognition, Control and Learning Algorithm Collaborative Research Chair (Science Tokyo). 

\bibliography{aaai2026}



\section*{Supplementary material}

# Geometry Meets Light: Leveraging Geometric Priors for Universal Photometric Stereo under Limited Multi-Illumination Cues
## ——Supplementary Material——

## 1. More Details of Our Training Dataset (PS-Perp)

As discussed in the main paper, unlike prior works that assume an orthographic projection, we adopt a perspective camera model with varying focal lengths. Apart from the camera settings, we follow the other configurations used in PS-Mix from SDM-UniPS. Therefore, we further detail the camera setup.

To simulate various levels of perspective distortion during rendering, we vary the camera's focal length $f$, randomly sampled from a specified range $[f_{\min}, f_{\max}]$ (i.e., 20mm-70mm or 20mm-1000mm). After determining the focal length, we calculated the distance between the camera and the scene by deriving the field of view (FoV) from the focal length, ensuring that the objects cover the frame of the image. As a result, the viewing directions at the edges of the image diverge more from the Z-axis, enhancing the perspective distortion of the scene.

To demonstrate the effects of perspective projection at different focal lengths, we present example renderings for three scenes at each of the following focal lengths: 15mm, 35mm, 70mm, 200mm, as well as under orthographic projection. In Fig. 1, we show three images (out of ten rendered per scene), along with their corresponding normal maps. Significant perspective distortion is evident at 15mm, with clear differences observed when compared to longer focal lengths.

## 2. Possible Q&A for Our Evaluation

Here, we provide answers to anticipated questions regarding our evaluation.

### 2.1. How PS-Perp and PS-Mix were mixed during training?

To ensure that the model learns to handle both orthographic and perspective projections, we train GeoUniPS on a mixed dataset consisting of PS-Mix from SDM-UniPS and our proposed PS-Perp dataset. PS-Mix provides scenes rendered under orthographic projection, while PS-Perp contains scenes rendered under perspective projection with varying focal lengths. Specifically, PS-Mix includes 34,921 orthographic scenes, and PS-Perp includes 60,297 perspective scenes, of which 44,220 are rendered with focal lengths shorter than 70mm. During training, we simply combine all the data and randomly sample batches from the shuffled dataset.

### 2.2. How images were captured in "Compariosn with 3D Scan"?

The object is placed less than 30cm from the camera, with a Sony Alpha 6400 mounted on a tripod. Illumination is provided by a handheld ring light, which is moved around the upper hemisphere of the object. We manually control the light source around the camera to ensure that four images exhibit very similar shading variations, thereby enhancing the limited multi-illumination cues. The room is kept dark to minimize ambient light and prevent it from influencing the reconstruction cues. The focal length varies as follows: Bear 33mm, Thinker 28mm, Box 47mm.

### 2.3. Detailed computational time?

The overall runtime of our model is 5.62 seconds per object on a single NVIDIA H100 GPU, using 16 input images at a spatial resolution of 512×512. This measurement, conducted on the DiLiGenT dataset, excludes I/O operations. Specifically, the encoder and decoder stages take 2.97 seconds and 2.52 seconds, respectively. When evaluated on the higher-resolution LUCES dataset (1024×1024), our model maintains efficient processing, with an average runtime of 22.77 seconds per object across 14 scenes.

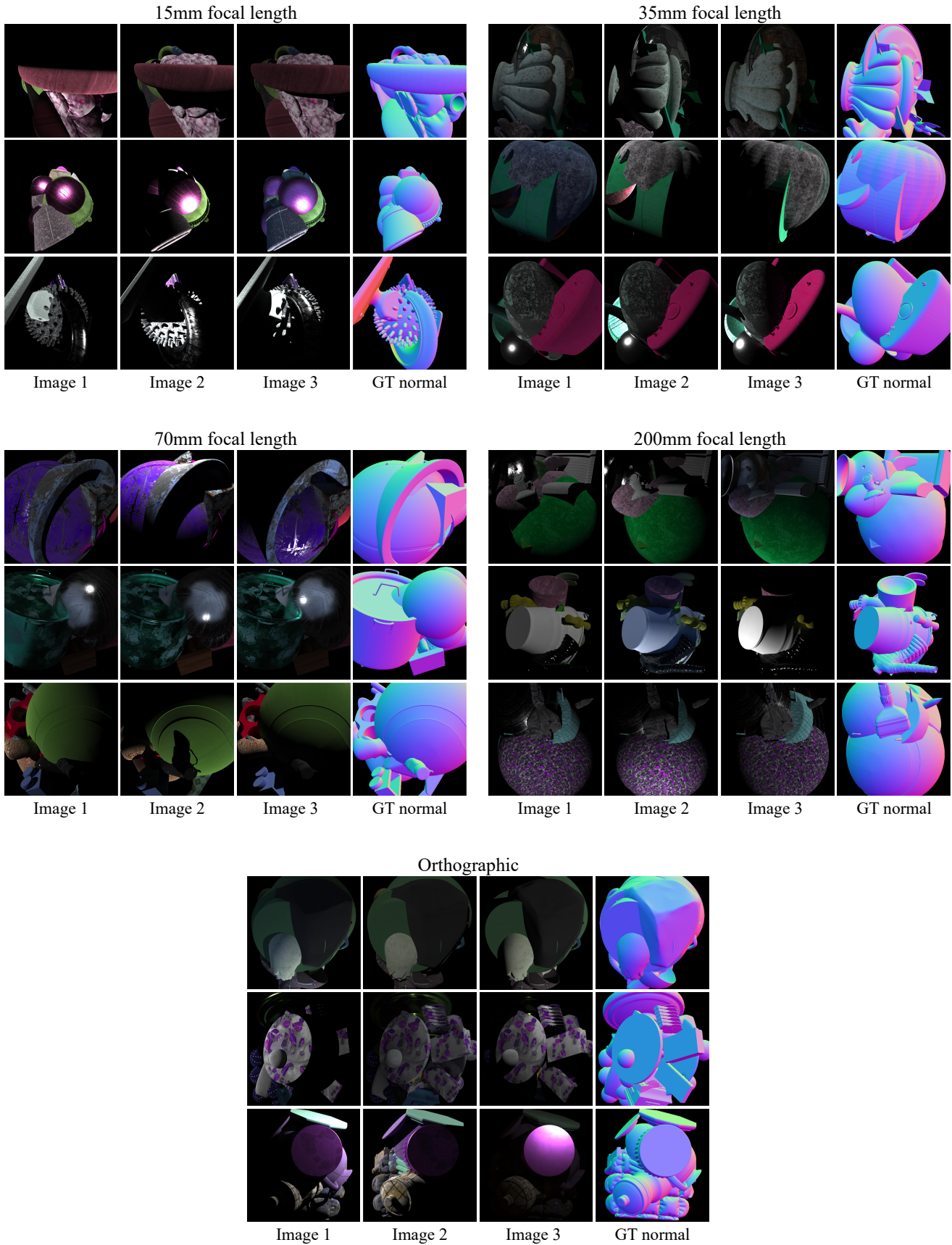

Figure 1. Examples of rendered scenes at different focal lengths (15mm, 35mm, 70mm, 200mm) and orthographic projection. Each example shows three out of ten rendered views per scene and their corresponding normal maps.

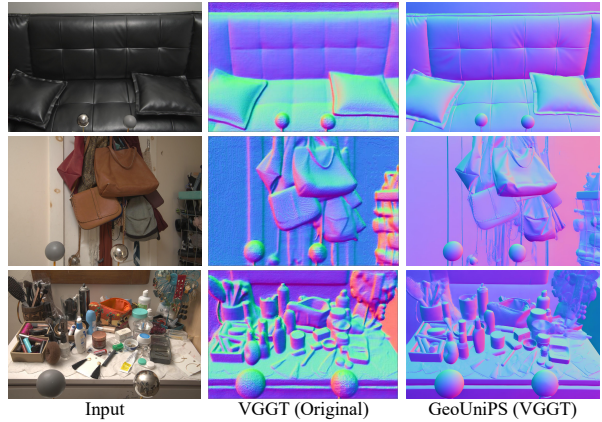

Figure 2. Comparison between our method using the VGGT backbone and the original VGGT depth-to-normal results.

### 2.4. How many model parameters?

GeoUniPS contains 1.27 billion parameters and consumes an average of 10.69GB of GPU memory during inference at 512×512 resolution. For 1024×1024 resolution, the memory usage increases slightly to 11.12GB on average. These results demonstrate that our model remains computationally efficient, even when scaling to higher resolutions while maintaining high prediction quality.

### 2.5. Implementation details of other methods (SDM-UniPS, Uni MS-PS and LINO-UniPS)?

For SDM-UniPS, Uni MS-PS, and LINO-UniPS, the code and pretrained models are publicly available, and we used them to perform all evaluations in our own environment. The method for computing normal errors is exactly the same, and any discrepancies from the results reported in the respective papers are due to these minor differences. We believe this approach offers the fairest basis for comparison.

### 2.6. Are multi-illumination cues really helpful?

A straightforward concern might be that the proposed method merely reproduces the output of the backbone, without leveraging meaningful contributions from multi-illumination cues. To address this, we computed surface normals directly from the depth map predicted by the pretrained VGGT.

As shown in Fig. 2, the normals reconstructed from the VGGT depth map lack fine detail, highlighting the effectiveness of incorporating multi-illumination cues in our dual-branch design.

## 3. Additional Analysis

Here, we provide additional analysis of the proposed method that could not be included in the main text due to space limitations.

### 3.1. Qualitative comparison between Encoder$_{Geo}$ w/ VGGT and Encoder$_{Geo}$ w/ MoGe from Table 4 (main paper)

In the main paper, we demonstrated GeoUniPS building upon VGGT, but Table 4 in the main paper presents results indicating that similar trends (though the performance was worse) are observed even when using other backbone (i.e., MoGe), suggesting that knowledge acquisition is still achievable. To qualitatively support this, we provide Fig. 3 comparing Encoder$_{Geo}$, w/ VGGT and Encoder$_{Geo}$, w/ MoGe from Table 4 (main). As shown here, these two variants produce very similar normal maps, validating that our approach is generalizable and not restricted to a specific backbone.

### 3.2. Comparison among different decoder designs

In the main paper, we claimed that "we find that the dual-scale decoder with MLP-based embedding provides the best trade-off between geometric detail and efficiency". To validate this, we provide a detailed comparison of four decoder variants as follow:

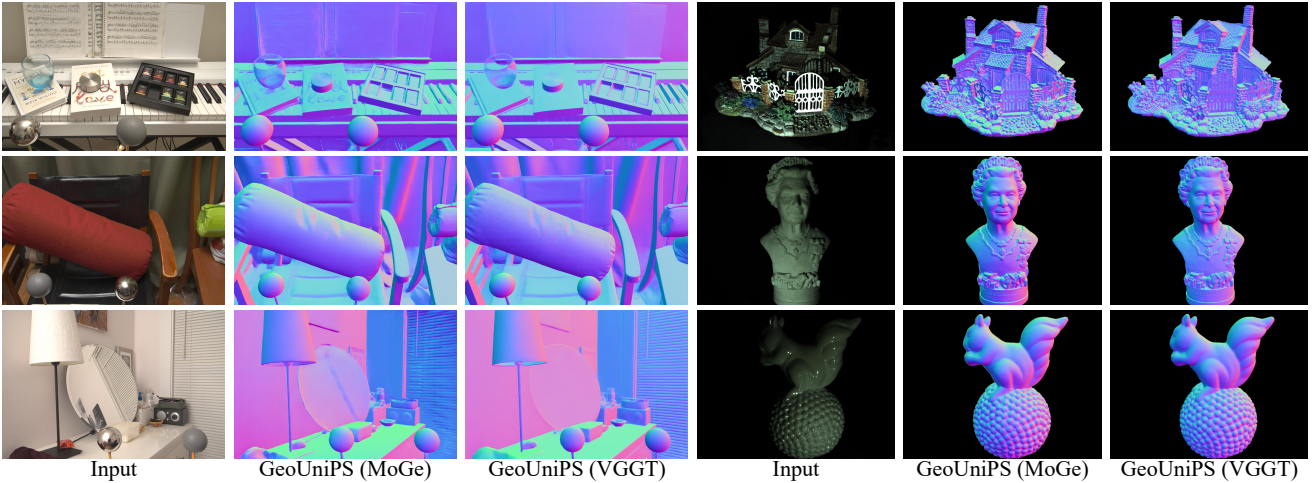

Figure 3. Comparison of our method using different geometric backbones. We compare GeoUniPS with VGGT and MoGe backbones. Our method with VGGT backbone achieves more detailed reconstructions, highlighting its effectiveness in capturing surface geometry.

Table 1. Ablation study on decoder configurations. MAE (↓) for different $K$ on DiLiGenT. All models use our dual-branch encoder.

| Decoder | DiLiGenT K=16 | K=96 |
|---|---|---|
| Decoder$_{single\text{-}scale}$ + Pixel-based Embed. | 5.22 | 4.91 |
| Decoder$_{dual\text{-}scale}$ + Pixel-based Embed. | 4.89 | 4.80 |
| Decoder$_{dual\text{-}scale}$ + Patch-based Embed. | 5.82 | NA |
| Decoder$_{dual\text{-}scale}$ + MLP-based Embed. | 4.77 | 4.65 |

1. a single-scale decoder using pixel-wise RGB input.

2. a dual-scale decoder using pixel-wise RGB input.

3. a dual-scale decoder with patch embedding.

4. a dual-scale decoder with pixel-wise MLP-based embedding.

The **pixel-based embedding** simply concatenates the pixel-wise RGB value with the feature vectors. The **patch-based embedding** is designed to encode local texture patterns. For each spatial location, a $21 \times 21$ RGB patch is extracted and flattened into a 1323-dimensional vector. This vector is first normalized with LayerNorm, then projected into a 128-dimensional latent space via a linear layer, followed by another LayerNorm for stabilization. This mechanism allows the decoder to access richer local texture and shading cues beyond individual pixels, thereby enhancing its ability to recover detailed surface normals. In contrast, the **MLP-based embedding** performs pixel-wise encoding. Each RGB pixel is first mapped to a 32-dimensional space via a linear layer and LeakyReLU activation, then further projected to a 256-dimensional embedding using another linear layer. Despite its simplicity and computational efficiency, this approach proves effective when combined with our dual-scale architecture, providing stable and accurate predictions while being more memory-efficient than patch embedding.

To evaluate these designs, we conduct both quantitative and qualitative experiments on the DiLiGenT datasets using K=16 and K=96 input images as shown in Table. 1 and Fig. 4. From the quantitative results, we observe that the dual-scale decoder consistently outperforms the single-scale variant, confirming its effectiveness in capturing multi-scale features. Additionally, while the patch embedding design performs well with a smaller number of input images, it suffers from excessive memory requirements as the input count increases. This supports our decision to favor the dual-scale decoder over the single-scale design. In the qualitative evaluation, while the patch embedding variant can enhance detail, it presents two major drawbacks: (1) extremely high memory consumption, and (2) increased noise in regions with significant depth variation, as previously discussed.

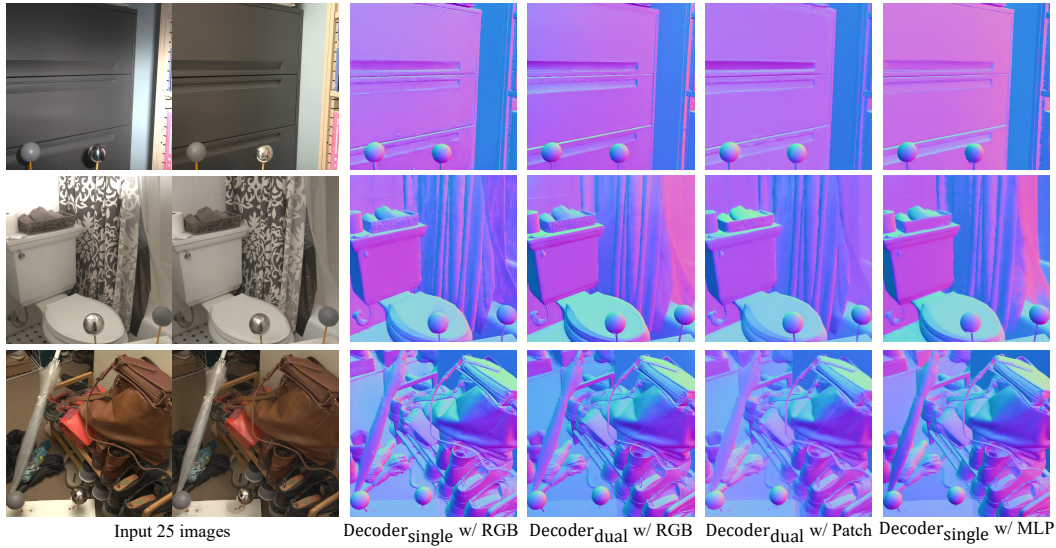

Figure 4. Comparison of different decoder designs. We present qualitative results from different decoder variants, including single-scale, dual-scale, and models with various embedding strategies; pixel-based (w/ RGB), ptach-based (w/ Patch) and MLP-based (w/MLP). Differences in detail can be observed across the designs.

## 4. Additional Results

As described in the main paper, we present additional results, including qulitative results and comparisons on DiLiGenT and LUCES datasets, more qualitative comparisons in Multi-illumination dataset.

### 4.1. Qualitative results for DiLiGenT and LUCES

To complement the quantitative results in the main paper, we provide additional qualitative comparisons on the DiLiGenT and LUCES datasets. While numerical metrics such as MAE offer an objective measure of performance, they may not fully reveal perceptual differences in surface normal quality. Visual comparisons allow for a more intuitive and direct understanding of how well each method captures fine-grained surface details, especially in challenging regions such as shadow boundaries, complex textures, and specular highlights.

Due to inherent randomness in training and testing procedures—such as weight initialization and data shuffling—model predictions may slightly vary across different runs. For consistency and clarity, we show representative results from one run in our visual comparisons. These examples are selected to fairly reflect the typical performance of each method.

To ensure a fair and comprehensive comparison, we also include the corresponding angular error maps for each method. These visualizations highlight regions with large estimation errors and help assess robustness under varying lighting conditions. All error maps are visualized with a fixed range from 0 to 30 degrees Mean Angular Error (MAE), allowing consistent comparison across different scenes and methods.

We present the results in Fig. 5 – Fig. 9.

### 4.2. More qulitative results from Multi-illumination dataset

To see more clearly how geometric prior contribute when multi-illumination cues are weak, we also demonstrate much more examples from Multi-illumination dataset. The results are shown in Fig. 10 – Fig. 13.

## 5. Failure Cases

Here, we present some failure cases of GeoUniPS. As shown, leveraging geometric priors from a pretrained 3D visual-geometry foundation model enables more robust and stable normal estimation, even when multi-illumination cues are limited.

However, as illustrated in Fig. 14, the method still struggles with transparent materials such as glass or clear mirrors. This limitation may arise from constraints inherent to the pretrained foundation model. We believe future work can address this issue by incorporating more advanced visual-geometry foundation models.

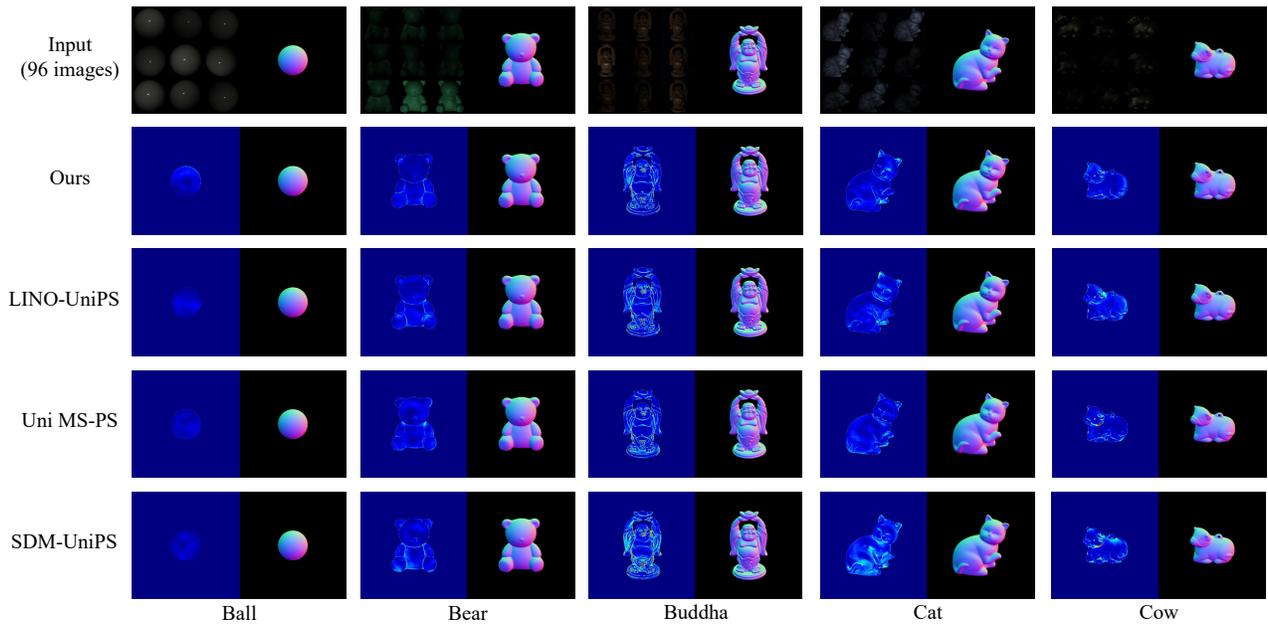

Figure 5. Qualitative results of different method on DiLiGenT dataset.

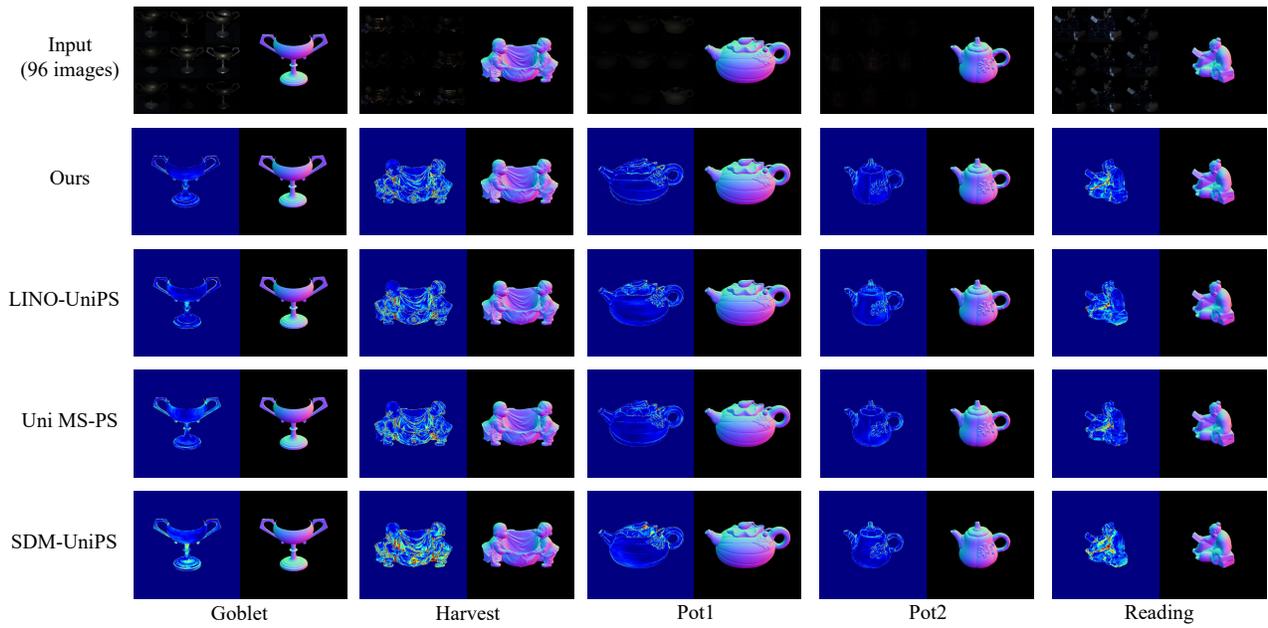

Figure 6. Qualitative results of different method on DiLiGenT dataset.

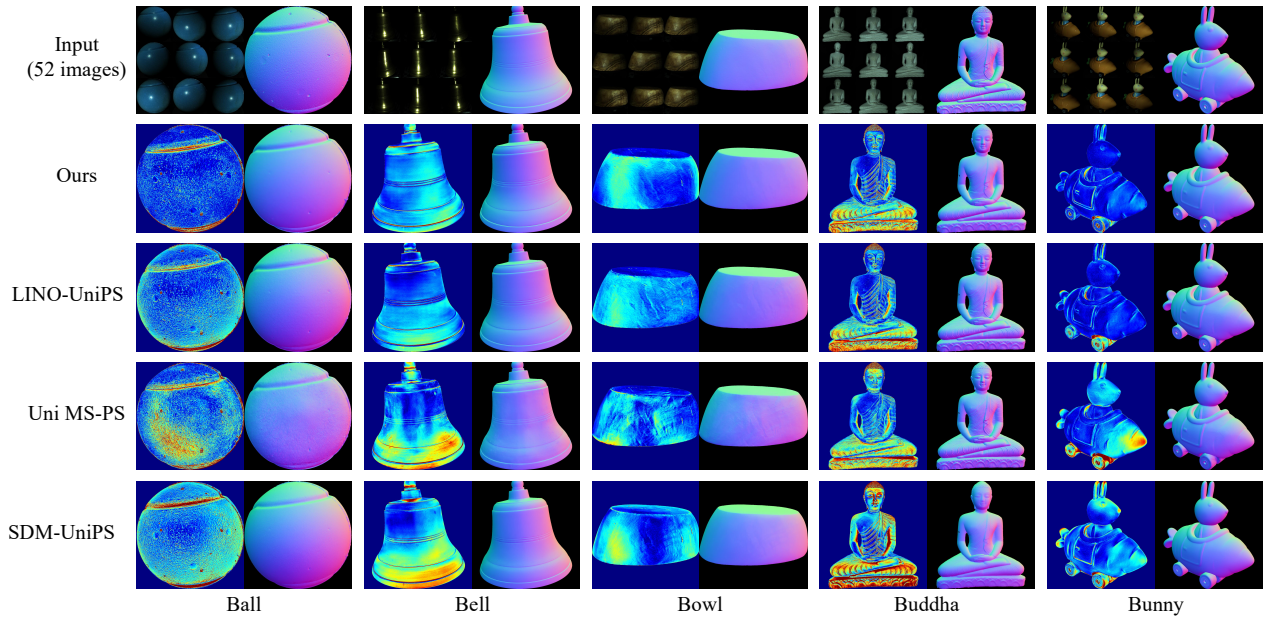

Figure 7. Qualitative results of different method on LUCES dataset.

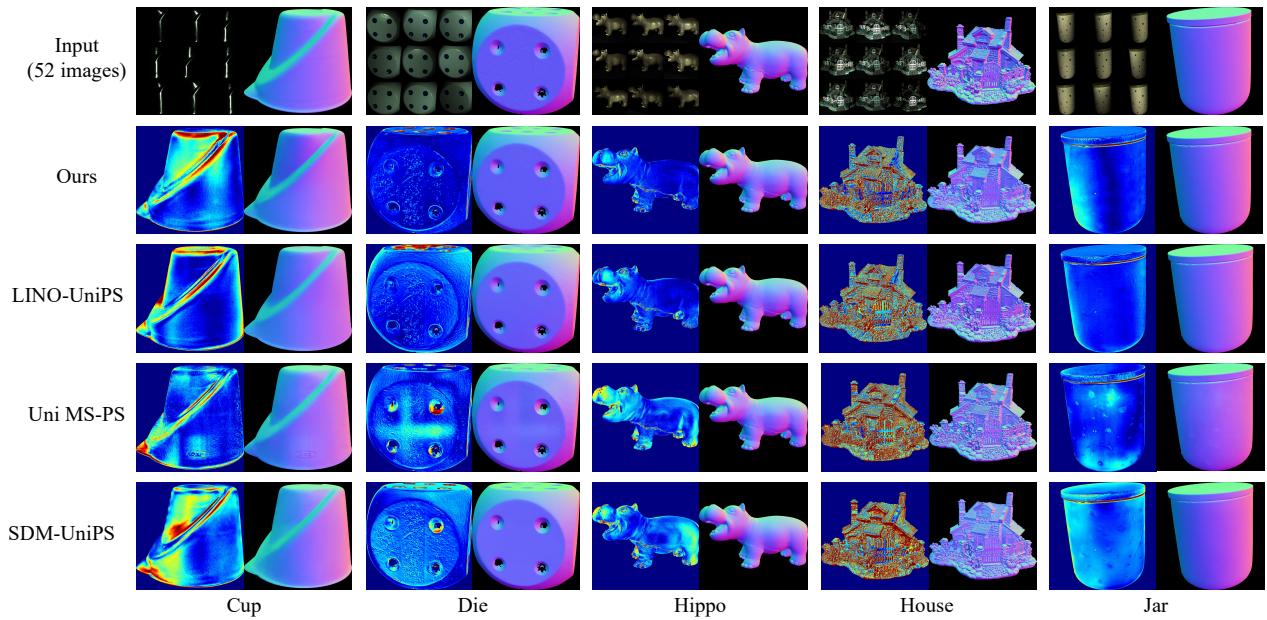

Figure 8. Qualitative results of different method on LUCES dataset.

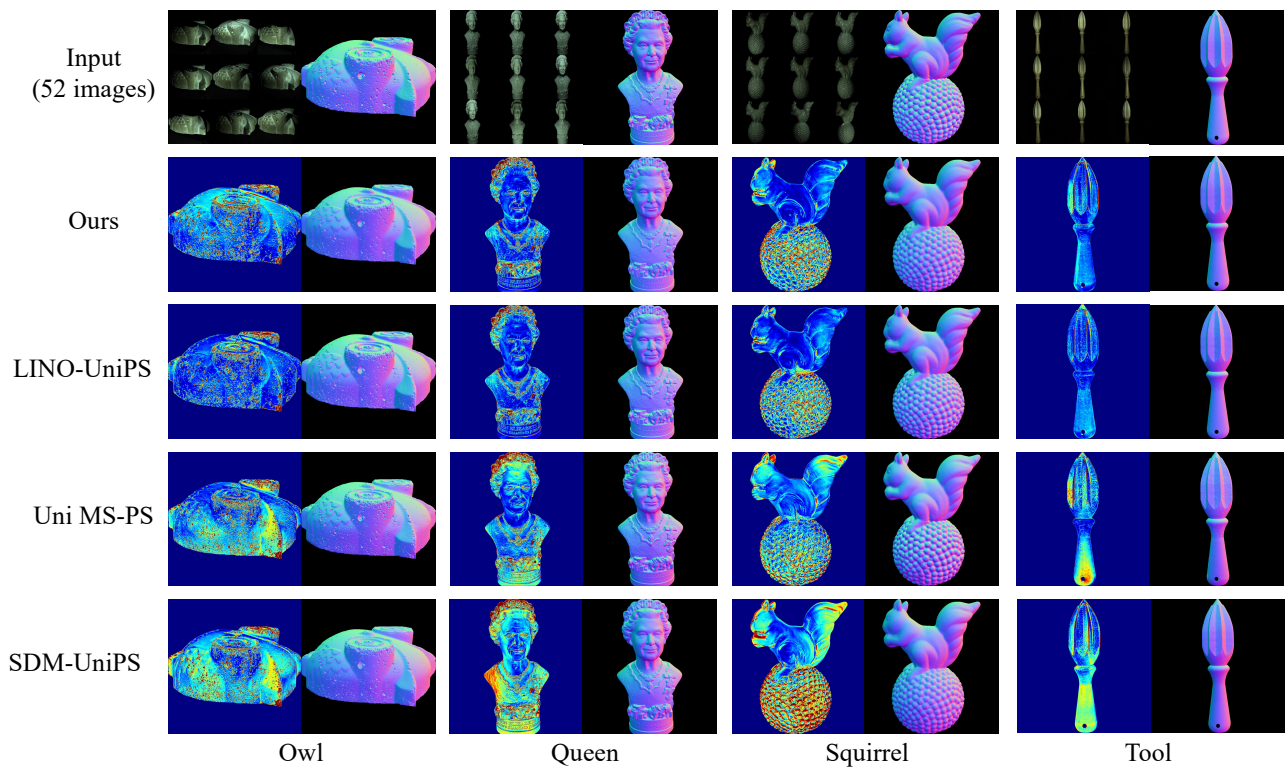

Figure 9. Qualitative results of different method on LUCES dataset.

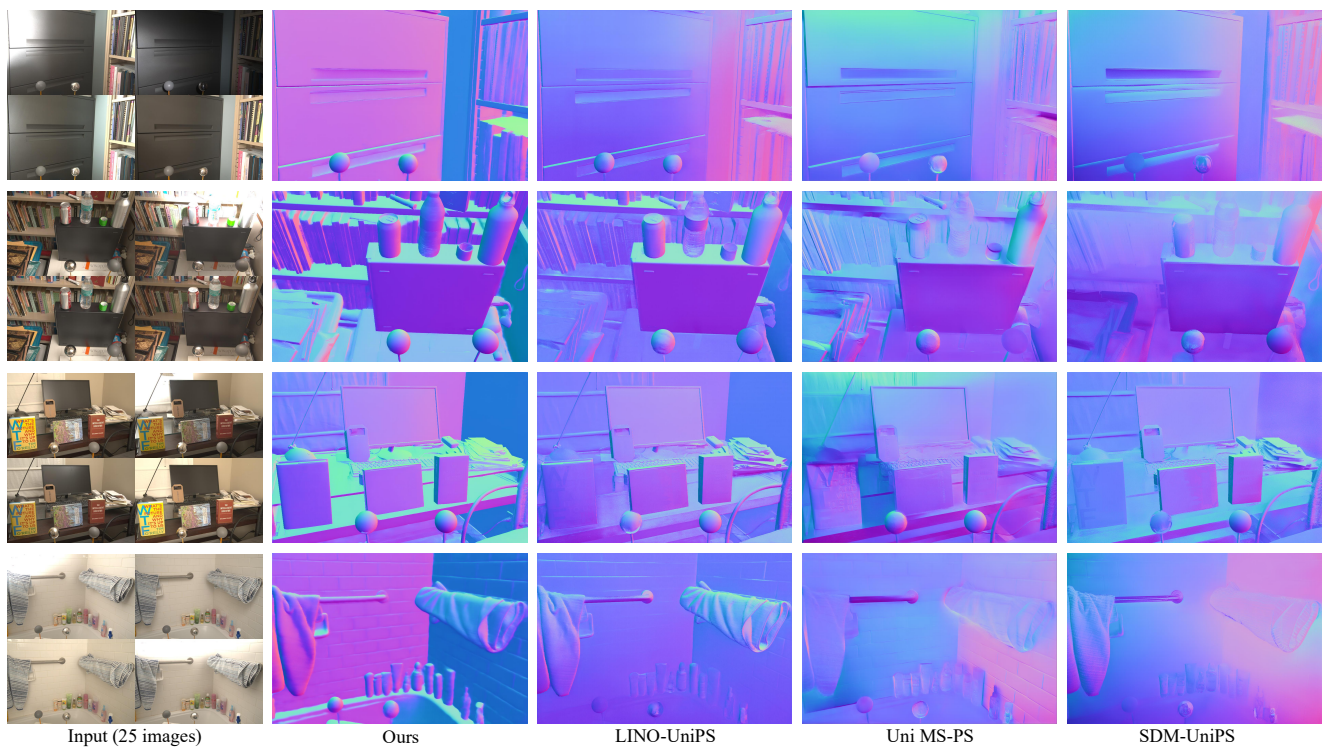

Figure 10. More qualitative results of different method on Multi-illumination dataset.

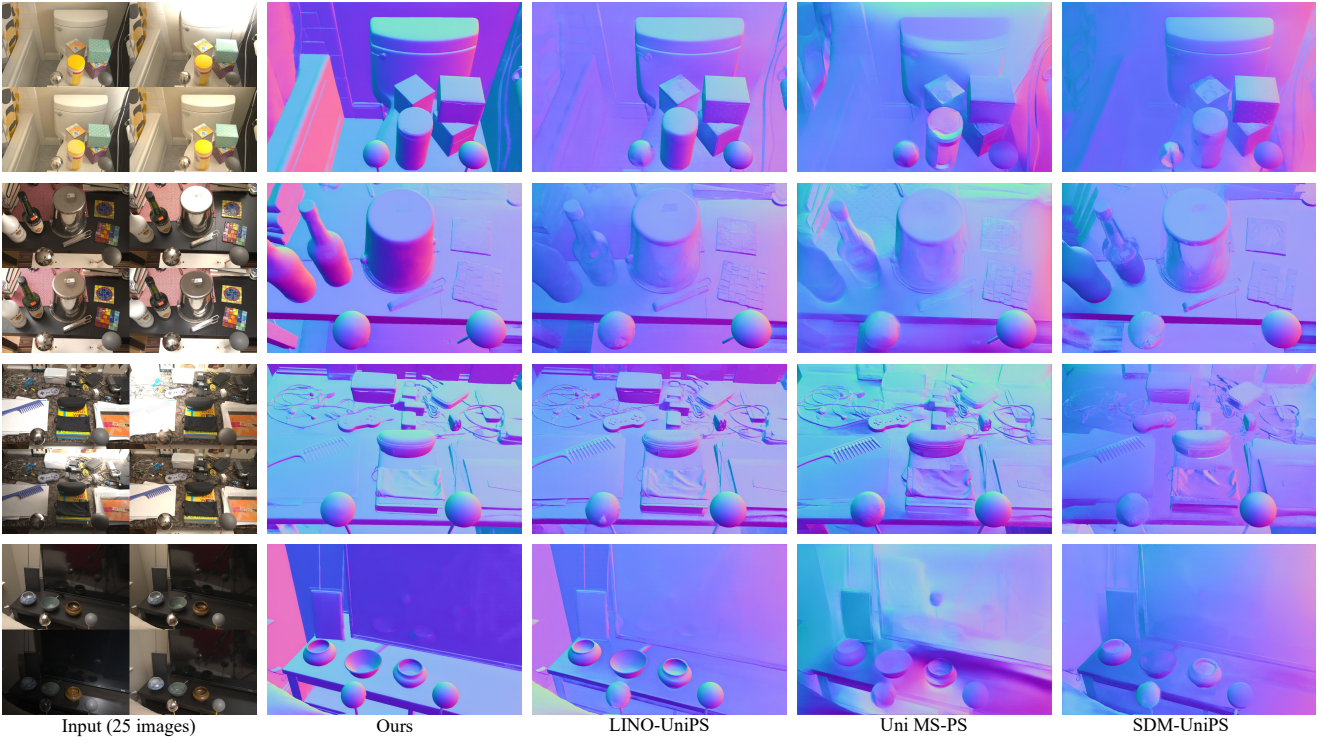

Input (25 images)    Ours    LINO-UniPS    Uni MS-PS    SDM-UniPS

Figure 11. More qualitative results of different method on Multi-illumination dataset.

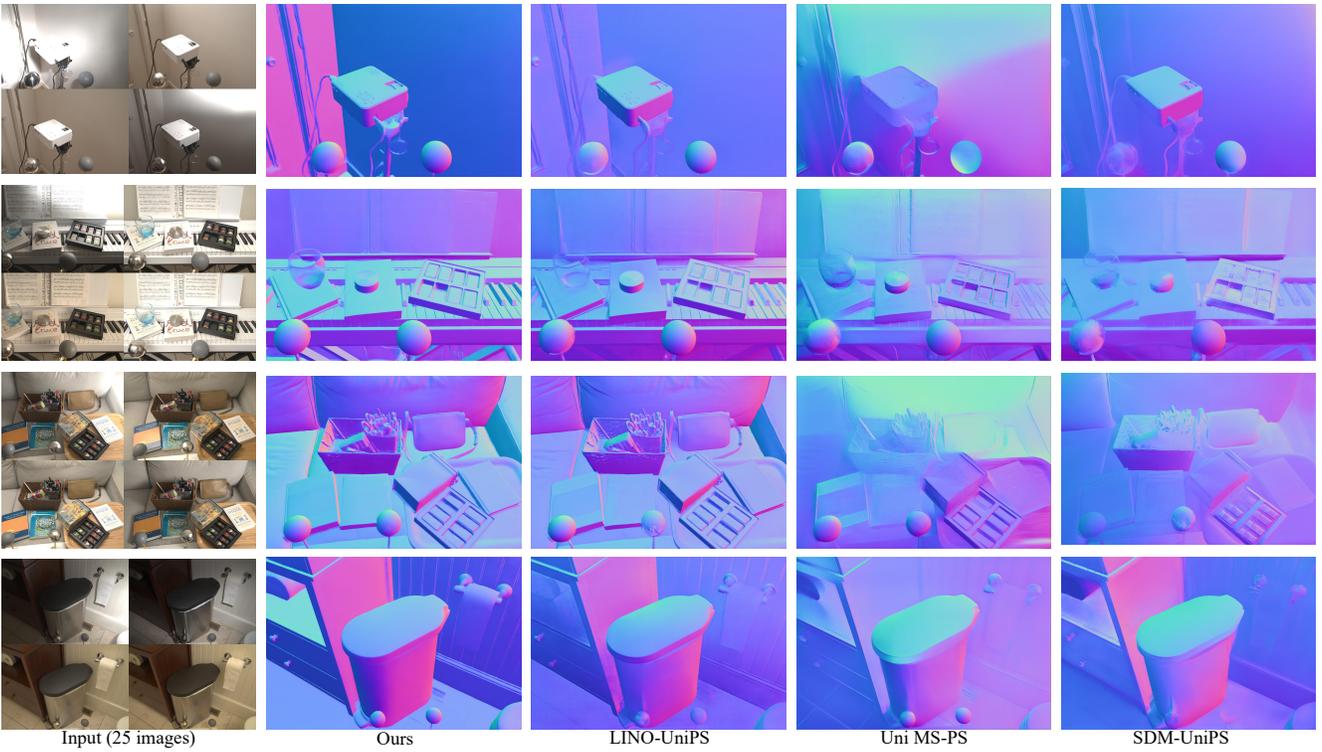

Input (25 images)    Ours    LINO-UniPS    Uni MS-PS    SDM-UniPS

Figure 12. More qualitative results of different method on Multi-illumination dataset.

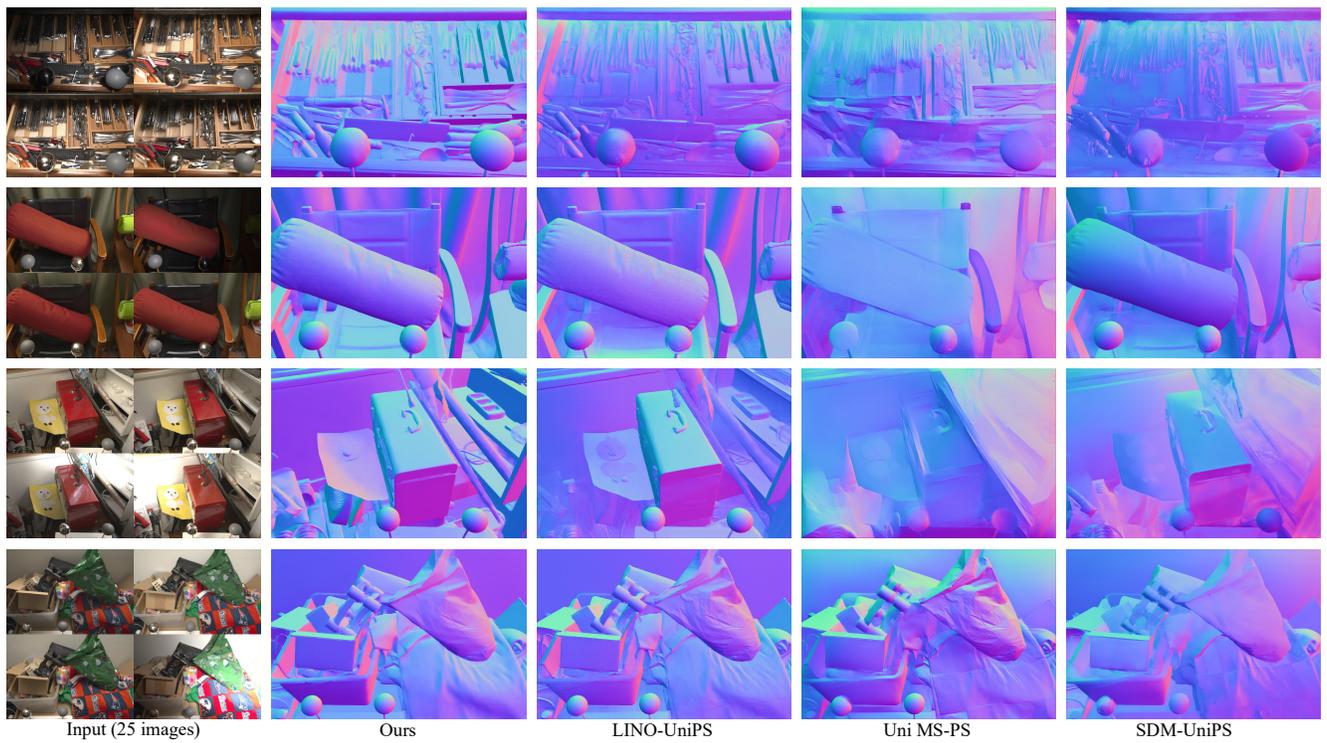

Figure 13. More qualitative results of different method on Multi-illumination dataset.

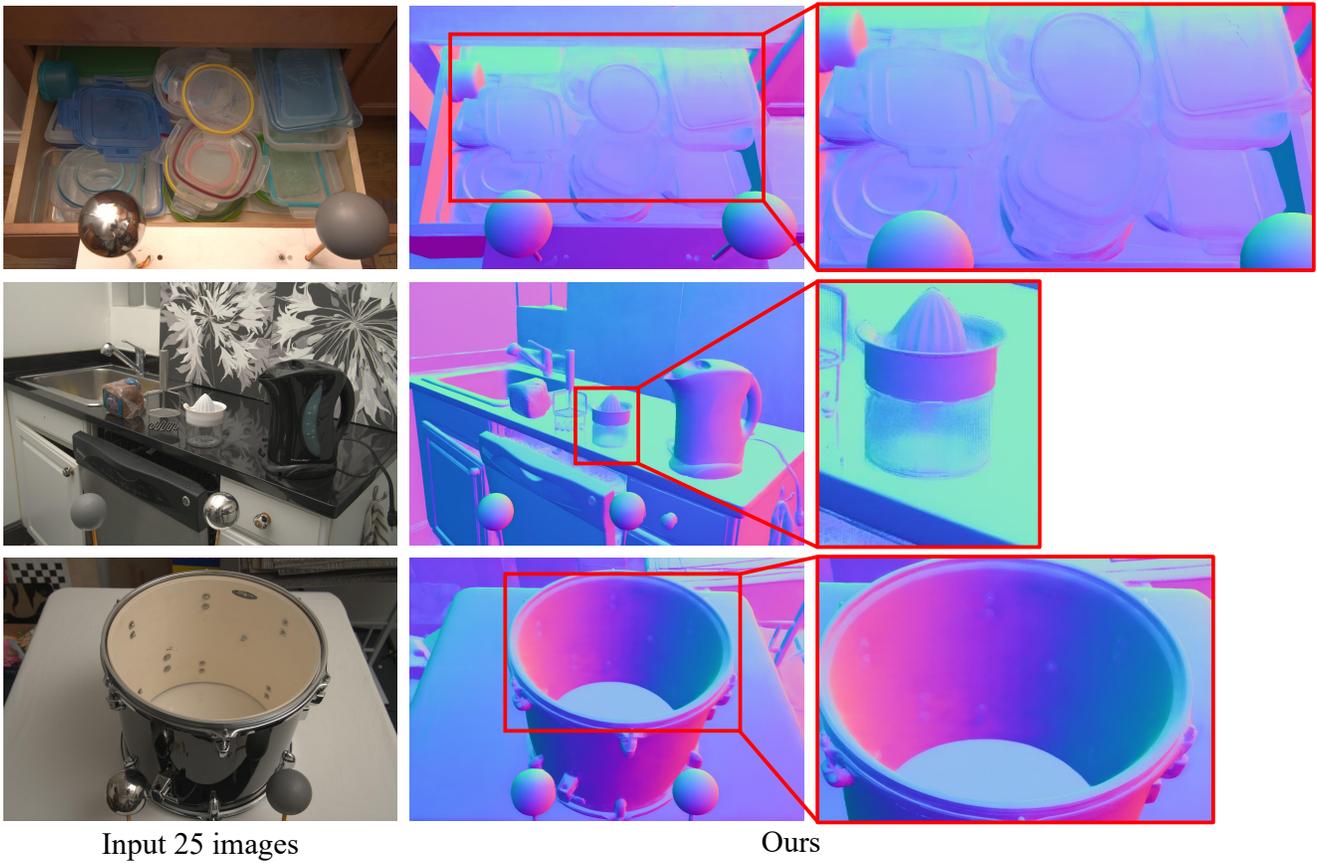

Input 25 images        Ours

Figure 14. Failure cases include inaccurate surface normal estimation on transparent or reflective objects (e.g., glass, mirrors).



\end{document}